\documentclass[letterpaper, 10 pt, conference]{ieeeconf}  
\IEEEoverridecommandlockouts
\makeatother
\usepackage{cite}
\usepackage{graphicx}
\usepackage{amsmath,amssymb,amsfonts}

\usepackage{comment}
\usepackage{picinpar}
\usepackage{amsmath}
\usepackage{url}
\usepackage{flushend}
\usepackage{colortbl}
\usepackage{soul}
\usepackage{booktabs}
\usepackage{multirow}
\usepackage{pifont}
\usepackage{color}
\usepackage{alltt}
\usepackage[hidelinks]{hyperref}
\usepackage{enumerate}
\usepackage{siunitx}
\usepackage{breakurl}
\usepackage{epstopdf}
\usepackage{pbox}
\usepackage{comment}
\usepackage{balance}

\usepackage{algorithm}
\usepackage{algorithmic}
\usepackage{amsfonts,mathrsfs}
\usepackage{graphicx}
\usepackage{subfigure}
\usepackage{bm}
\let\labelindent\undefined
\usepackage{enumitem}
\usepackage[utf8]{inputenc}
\usepackage{tikz}

\newcommand{\be}{\begin{equation}}
\newcommand{\ee}{\end{equation}}
\newcommand{\bea}{\begin{eqnarray}}
\newcommand{\eea}{\end{eqnarray}}
\newcommand{\ba}{\begin{array}}
\newcommand{\ea}{\end{array}}

\def\BibTeX{{\rm B\kern-.05em{\sc i\kern-.025em b}\kern-.08em
    T\kern-.1667em\lower.7ex\hbox{E}\kern-.125emX}}

\DeclareRobustCommand{\TableCapWord}[2]{%
  {\upshape\fontsize{8pt}{9.6pt}\selectfont\MakeUppercase{#1}}%
  {\upshape\fontsize{6.4pt}{9.6pt}\selectfont\MakeUppercase{#2}}}
\DeclareRobustCommand{\TableCapMinor}[1]{%
  {\upshape\fontsize{6.4pt}{9.6pt}\selectfont\MakeUppercase{#1}}}

\begin{document}

\title{\textbf{Detecting Heel Strike and toe off Events Using Kinematic Methods and LSTM Models}}
\author{Longbin~Zhang, ZhiZhang Li\textsuperscript{*}, XinYi Fu, Yi Xie, ZeKun Liu, Xiaoyue Yan, Suiyuan Wang, Te Zhang, \\Hui Zhang, Kailun Yang,  Tsung-Lin Wu,  Prayook Jatesiktat, Ananda Sidarta, and Wei Tech Ang
\thanks{*This work was supported by the National Natural Science Foundation of China under Grant 62503165 and 62673192, Hunan Provincial Research and Development Project under Grant 2026QK3021, Yuelushan Center for Industrial Innovation, and Health Commission of Hunan Province. The authors also acknowledge the support from Nanyang Technological University, the Agency for Science Technology and Research (A*STAR), the National Healthcare Group (NHG). (Corresponding author: ZhiZhang Li; e-mail: zhizhangli@hnu.edu.cn)}
\thanks{L. Zhang is with the School of Artificial Intelligence and Robotics, Hunan University, Changsha 410082, China, and the School of Mechanical and Aerospace Engineering, Nanyang Technological University, Singapore 639798 (e-mail: longbin@hnu.edu.cn).}
\thanks{Z. Li, X. Fu, Y. Xie, Z. Liu, H. Zhang, and K. Yang are with the School of Artificial Intelligence and Robotics, Hunan University, Changsha 410082, China (e-mail: zhizhangli@hnu.edu.cn; foxn@hnu.edu.cn; xieyi@hnu.edu.cn; liuzekun@hnu.edu.cn; zhanghui1983@hnu.edu.cn; kailun.yang@hnu.edu.cn, respectively).}
\thanks{X. Yan, T. Wu, P. Jatesiktat, and W. T. Ang are with the School of Mechanical and Aerospace Engineering, Nanyang Technological University, Singapore 639798 (e-mail: xiaoyue001@e.ntu.edu.sg; tsunglin.wu@ntu.edu.sg; prayook001@e.ntu.edu.sg; WTAng@ntu.edu.sg, respectively).}
\thanks{S. Wang is with the Fourth Hospital of Changsha, Hunan Normal University, Changsha 410006, China (e-mail: wangsuiyuan@hunnu.edu.cn).}
\thanks{T. Zhang is with the Department of Stomatology, Hunan Provincial People's Hospital, Hunan Normal University, Changsha 410005, China (e-mail: 286414895@qq.com).}
\thanks{A. Sidarta is with the Rehabilitation Research Institute of Singapore, Lee Kong Chian School of Medicine, Nanyang Technological University, Singapore 308232 (e-mail: ananda.sidarta@ntu.edu.sg).}
}

\maketitle
\thispagestyle{empty}
\pagestyle{empty}
\begin{abstract}
Accurate gait event detection is crucial for gait analysis, rehabilitation, and assistive technology, particularly in exoskeleton control, where precise identification of stance and swing phases is essential. This study evaluated the performance of seven kinematics-based methods and a Long Short-Term Memory (LSTM) model for detecting heel strike and toe-off events across 4363 gait cycles from 588 able-bodied subjects. The results indicated that while the Zeni et al. method achieved the highest accuracy among kinematics-based approaches, other methods exhibited systematic biases or required dataset-specific tuning. The LSTM model performed comparably to Zeni et al., providing a data-driven alternative without systematic bias. These findings highlight the potential of deep learning-based approaches for gait event detection while emphasizing the need for further validation in clinical populations and across diverse gait conditions. Future research will explore the generalizability of these methods in pathological populations, such as individuals with post-stroke conditions and knee osteoarthritis, as well as their robustness across varied gait conditions and data collection settings to enhance their applicability in rehabilitation and exoskeleton control.
\end{abstract}
\IEEEpeerreviewmaketitle
\section[INTRODUCTION]{{\fontsize{10pt}{12pt}\selectfont INTRODUCTION}}
Exoskeletons are emerging as powerful assistive technologies designed to enhance mobility, restore gait function, and reduce the physical burden on users in both rehabilitation and everyday settings \cite{li2018physical,zhang2022ankle}. A key challenge in exoskeleton control is the precise timing of assistance, which relies on accurate gait event detection to differentiate between stance and swing phases \cite{li2019hybrid,luo2022trajectory}.
Misidentification of these events can lead to improper assistance timing, reducing efficiency, increasing metabolic cost, and potentially compromising user safety. Gait event detection is, therefore, a fundamental component of gait analysis, ensuring the precise identification of key moments within the gait cycle, such as heel strike (HS) and toe off (TO) \cite{kim2022deep}. Accurate detection is essential for calculating spatiotemporal parameters and kinematic variables, which are critical for optimizing exoskeleton control strategies. Errors in gait event identification can propagate through the system, leading to inaccurate joint angle estimations and impaired motion synchronization \cite{dumphart2023robust, zeni2008two}. In clinical settings, precise gait event detection is vital for diagnosing gait abnormalities, monitoring rehabilitation progress, and tailoring interventions \cite{baker2013measuring}. It also plays a crucial role in providing real-time feedback to patients, preventing secondary complications, and customizing assistive devices \cite{simon2004quantification}. Beyond clinical applications, accurate gait event detection supports the optimization of prosthetics and orthotics, enhances athletic performance, and enables predictive modeling for fall risk \cite{sutherland2002evolution}. Given its significance in both rehabilitation and assistive robotics \cite{zhang2021modeling}, improving gait event detection is essential for advancing exoskeleton technology, ensuring smooth user-exoskeleton interaction, and enhancing mobility for individuals with gait impairments.

The gold standard for identifying HS and TO in gait analysis involves using force plates with predefined vertical ground reaction force (GRF) thresholds, typically ranging from 5 to 20 N 
\cite{zahradka2020evaluation,tirosh2003identifying}, 
in a laboratory setup where subjects walk across a walkway embedded with force plates \cite{bruening2014automated, lempereur2020new, bonci2022algorithm}. 
However, the availability of force plates and the ability to achieve sufficient clean foot contacts can be significant challenges, particularly in cases with pathological characteristics such as foot drag, small step size, or the use of walking aids \cite{dumphart2023robust}. Additionally, employing force plates often necessitates a force-sensitive walkway or multiple force plates, which are expensive and impractical, requiring each footstep to occur on a single force plate and ensuring only one foot contacts the plate at a time \cite{hansen2002simple}. Portable alternatives, such as pressure insoles or foot switches, offer some flexibility but require additional equipment and modifications to the subject’s footwear, limiting their use in populations with abnormal gait patterns, especially those lacking sufficient foot clearance during the swing phase. Furthermore, the accuracy of these portable methods heavily depends on the precise placement of sensors \cite{zeni2008two}. Despite the challenges, force plate thresholding remains the gold standard, but there is a continuous need for advancements in gait event detection methods to improve accuracy and applicability across diverse clinical and research settings.

Kinematics-based models and deep learning approaches are widely used for gait event detection in both healthy and pathological populations \cite{zeni2008two,desailly2009foot,ghoussayni2004assessment,o2007automatic,hreljac2000algorithms, lempereur2020new, kim2022deep}.
These models often rely on foot or pelvis segments captured through motion capture systems or wearable sensors like IMUs \cite{french2020comparison, voisard2024automatic, bonci2022algorithm}. French et al. \cite{french2020comparison} found that horizontal ankle-heel distance was most accurate for detecting heel strike during overground walking, while sacral-heel distance was best for toe off (TO). Voisard et al. \cite{voisard2024automatic} used inertial measurement units (IMUs) to detect gait events, showing high accuracy in healthy subjects (TO: 8 ms, HS: 7 ms) and reasonable accuracy in multiple sclerosis patients (TO: 23 ms, HS: 15 ms). Bonci et al. \cite{bonci2022algorithm} combined marker position and velocity, achieving over 99\% sensitivity but limited by small sample sizes and marker requirements.
Deep learning models, including Long Short-Term Memory (LSTM) networks, have emerged as alternatives. Lempereur et al. \cite{lempereur2020new} used LSTM to detect gait events in children with gait disorders, achieving high accuracy with HS detected within 5.5 ms and TO within 10.7 ms. Kim et al. \cite{kim2022deep} achieved 89.7\% detection for HS and 71.6\% for TO in children with cerebral palsy. Filtjens et al. \cite{filtjens2020data} employed Temporal Convolutional Networks for Parkinson's disease, achieving F1-scores exceeding 0.99. Dumphart et al. \cite{dumphart2023robust} used LSTM to detect events in both healthy individuals and those with pathologies, reporting high detection rates but emphasizing the influence of marker placement and variability in laboratory setup.
Despite advances, few studies have compared kinematics-based and deep learning models across data centers and diverse subject groups.

The objective of this study was to detect heel strike and toe off events using kinematics-based methods and an LSTM model and to evaluate their prediction accuracy. We hypothesized that the LSTM model would achieve comparable or superior accuracy to traditional kinematics-based methods while maintaining consistency across trials.

\begin{figure*}[!]
	\centering
	\includegraphics[width=5.2in]{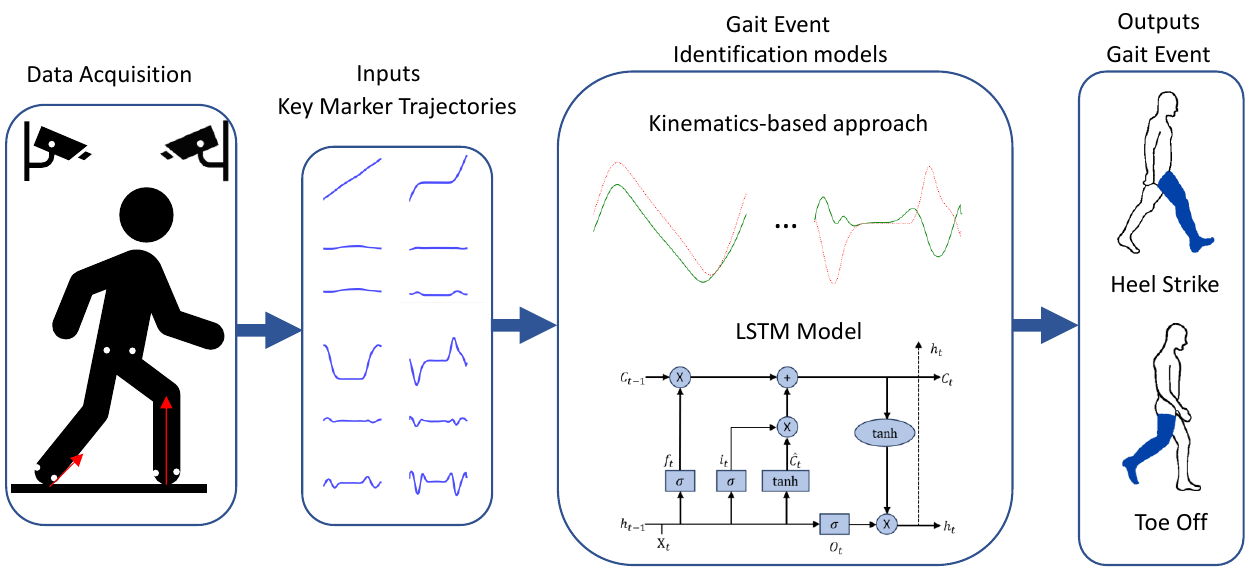}
	\caption{Schematic diagram of gait event identification using kinematics-based or machine learning models. The inputs for these models consist of key marker trajectories from the toe, heel, or pelvis, acquired and post-processed using a 3D motion capture system.
 }
\label{Overview}
\end{figure*}

\section[METHODS]{{\fontsize{10pt}{12pt}\selectfont METHODS}}

\subsection{Experimental Setup}
A total of 4363 gait cycles from 588 able-bodied subjects (age: 45 $\pm$ 16.2 years; sex: 317F/271M; weight: 62.7 $\pm$ 13.5 kg; height: 164.4 $\pm$ 9.1 cm) were collected from the Rehabilitation Research Institute of Singapore. The study was approved by the Nanyang Technological University Institutional Review Board (IRB-2022-382). Marker trajectories were recorded using a marker-based Miqus M3 Qualisys system (200 Hz) in Singapore. Ground reaction forces were simultaneously measured and used to identify heel strike and toe off events using a 20 N threshold, serving as the ground truth for both kinematics-based and machine learning models (Fig. \ref{Overview}). 
\begin{table*}[!]
\centering	
\caption{\TableCapMinor{The} \TableCapWord{K}{inematics-}\TableCapWord{B}{ased} \TableCapWord{A}{pproaches} \TableCapWord{U}{sed} \TableCapMinor{to} \TableCapWord{D}{etect} \TableCapWord{HS}{} \TableCapMinor{and} \TableCapWord{TO}{}}
\renewcommand{\arraystretch}{1.5}
\begin{tabular}{m{11em} m{6em} m{20em} m{20em}} 
\hline\hline
\multicolumn{1}{c}{\textbf{Method}} & \multicolumn{1}{c}{\textbf{Markers}} & \multicolumn{1}{c}{\textbf{HS Detection Features}} & \multicolumn{1}{c}{\textbf{TO Detection Features}} \\
\hline\hline
Foot trajectories relative to pelvis
(Zeni et al. 2008)& Heel, toe, and pelvis & Local maximum horizontal position of the heel marker relative to the pelvis  &  Local minimal horizontal position of the toe marker relative to the pelvis  \\
\hline
High-pass filtered foot markers 
(Desailly et al. 2009) & Heel and toe &  First maximum between high pass filtered (0.5 * gait frequency) horizontal heel marker trajectory &  Last minimum between high pass filtered (1.1 * gait frequency) horizontal toe marker trajectory \\
\hline
Foot markers velocity extreme (O'Connor et al. 2007) & Heel and toe &  Local minimal vertical velocity of the mid-point of heel and toe markers & Local maximum vertical velocity of the mid-point of heel and toe markers\\
\hline
Foot markers Velocity Threshold 
(Ghoussayni et al. 2004) & Heel and toe & Sagittal velocity of heel marker lower than 50 cm/s  &  Sagittal velocity of toe marker higher than 50 cm/s 
\\
\hline
Foot markers acceleration extreme (Hreljac et al. 2000) & Heel and toe & Local maximum vertical acceleration of the heel marker, with the condition that the derivative of the acceleration (jerk) was zero. & Local maximum horizontal acceleration of the toe marker, also with the constraint that the relevant jerk was zero\\
\hline
Foot markers acceleration extreme (Hsue et al. 2007) & Heel and toe & Local minimal horizontal acceleration of heel marker   &  Local maximum horizontal acceleration of toe marker  \\
\hline
Position and Velocity combined
(Bonci et al. 2022) & Heel, Toe and Pelvis & Initially detected using the method from Zeni et al. (2008) and then refined based on the 3D velocity of heel marker: lower than 0.5 * walking speed for rearfoot contacts and 0.8 * walking speed for forefoot contacts  &  Initially detected using the method from Zeni et al. and then refined when the 3D velocity of the toe marker exceeded 0.8 times the walking speed, and further refined when the 3D velocity of the heel marker decreased after its local peak\\
\hline\hline 
\end{tabular}\label{Kinematics}
\end{table*}

\subsection{Gait Event Identification Models}
\textit{(a) Kinematics-based Models}

To identify heel strike and toe off gait events, we evaluated seven kinematics-based methods derived from peer-reviewed literature. These methods employed different biomechanical principles, utilizing positional, velocity, or acceleration-based criteria applied to key anatomical markers, including those placed on the heel, toe, and pelvis (Table \ref{Kinematics}). Each approach had unique advantages and limitations, particularly in terms of sensitivity to gait variations and robustness across different walking conditions.
\begin{itemize}
    \item Positional-Based Methods:
\end{itemize}

\textbf{Zeni et al.} (2008) proposed a widely used method for detecting HS and TO based on the local extrema of heel and toe marker positions relative to the pelvis. Specifically, HS was identified when the heel marker reached its maximum horizontal displacement, while TO was detected when the toe marker reached its minimum horizontal displacement, both measured relative to the pelvis (Table \ref{Kinematics}).

\textbf{Desailly et al.} (2009)  introduced a filtering-based approach in which high-pass filtering was applied to horizontal heel and toe marker trajectories to eliminate low-frequency drift. HS was detected at the first maximum in the high-pass filtered (0.5 * gait frequency) horizontal heel marker trajectory, while TO was identified at the last minimum in the high-pass filtered (1.1 * gait frequency) horizontal toe marker trajectory. This method was designed to enhance accuracy by mitigating the effects of marker noise and soft tissue artifacts.

\begin{itemize}
    \item Velocity-Based Methods:
\end{itemize}

\textbf{O'Connor et al.} (2007) employed a velocity-based detection method, defining HS and TO based on vertical velocity extrema. Specifically, HS was identified at the local minimum and TO at the local maximum of the vertical velocity of the midpoint between the heel and toe markers.

\textbf{Ghoussayni et al.} (2004) introduced a threshold-based velocity approach, where HS was detected when the sagittal velocity of the heel marker fell below 5 cm/s, and TO was identified when the sagittal velocity of the toe marker exceeded 5 cm/s. However, the velocity threshold recommended by Ghoussayni et al. was later found to be too low and was increased to 50 cm/s by Bruening et al. \cite{bruening2014automated}. In this study, we adopt the revised threshold of 50 cm/s.

\begin{itemize}
    \item Acceleration-Based Methods:
\end{itemize}

\textbf{Hreljac et al.} (2000) used acceleration profiles to detect HS and TO, incorporating vertical and horizontal acceleration components to capture gait event dynamics. HS was identified at the local maximum vertical acceleration of the heel marker, provided that the derivative of acceleration (jerk) was zero. Similarly, TO was detected at the local maximum horizontal acceleration of the toe marker, also under the condition that the corresponding jerk was zero.

\textbf{Hsue et al.} (2007) extended acceleration-based detection by introducing jerk constraints (rate of change of acceleration). HS was identified at the local minimum of the heel marker's horizontal acceleration, while TO was detected at the local maximum of the toe marker's horizontal acceleration.

\begin{itemize}
    \item Hybrid Approaches:
\end{itemize}

\textbf{Bonci et al.} (2022) refined Zeni’s approach by incorporating 3D velocity thresholds relative to walking speed. HS was initially detected using the method from Zeni et al. (2008) and then adjusted based on the 3D velocity of the heel marker: below 0.5 * walking speed for rearfoot contacts and 0.8 * walking speed for forefoot contacts. TO was first identified using Zeni et al.’s (2008) method, then refined when the 3D velocity of the toe marker exceeded 0.8 * walking speed, with a further adjustment when the 3D velocity of the heel marker decreased after reaching its local peak.

\begin{figure*}[!]
	\centering
	\includegraphics[width=6in]{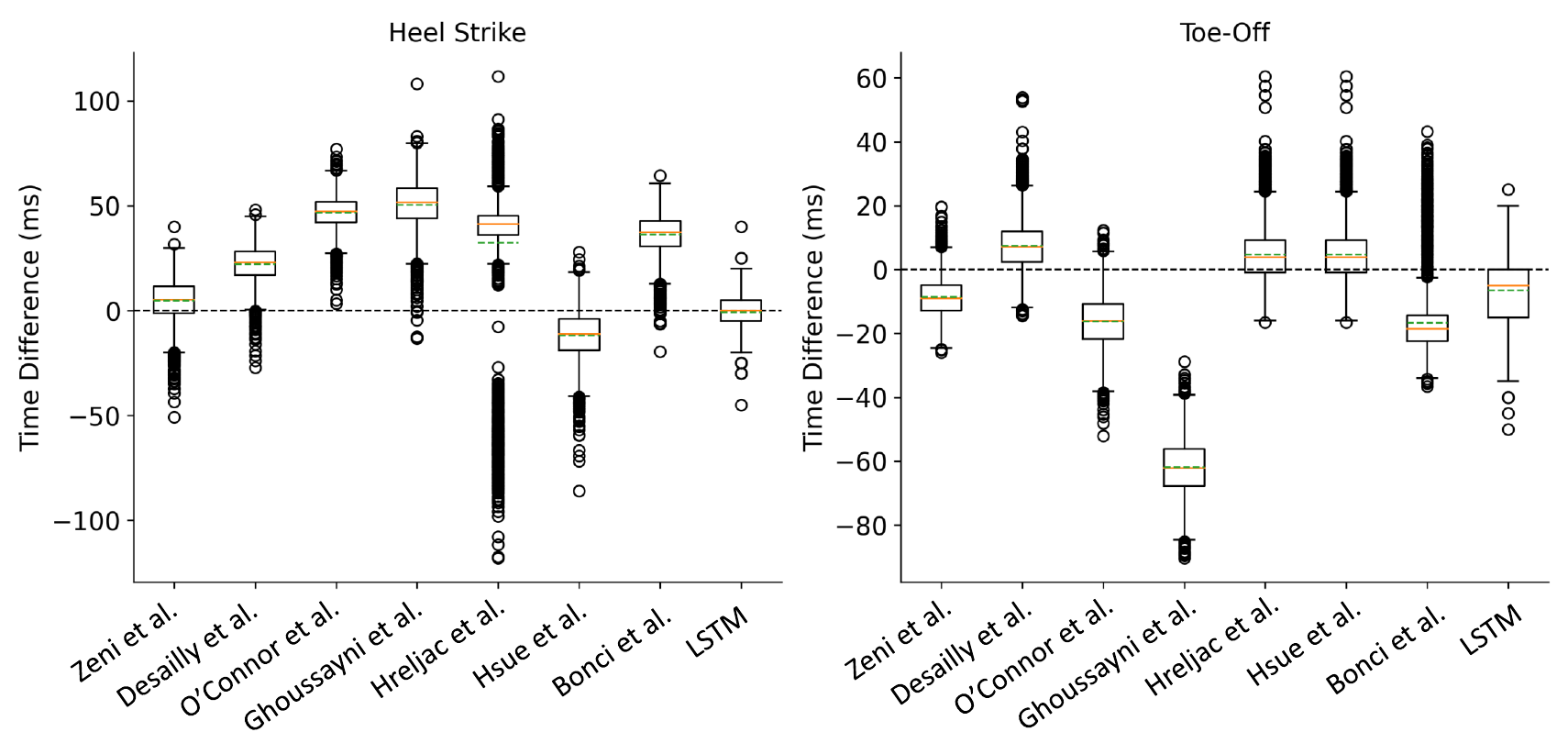}
	\caption{The boxplot of prediction errors (time differences) between the measured and calculated heel strike (left) and toe off (right) events using various kinematics-based methods and the LSTM model.
 }
\label{boxplot}
\end{figure*}

\begin{figure}[!]
	\centering
	\includegraphics[width=3.4in]{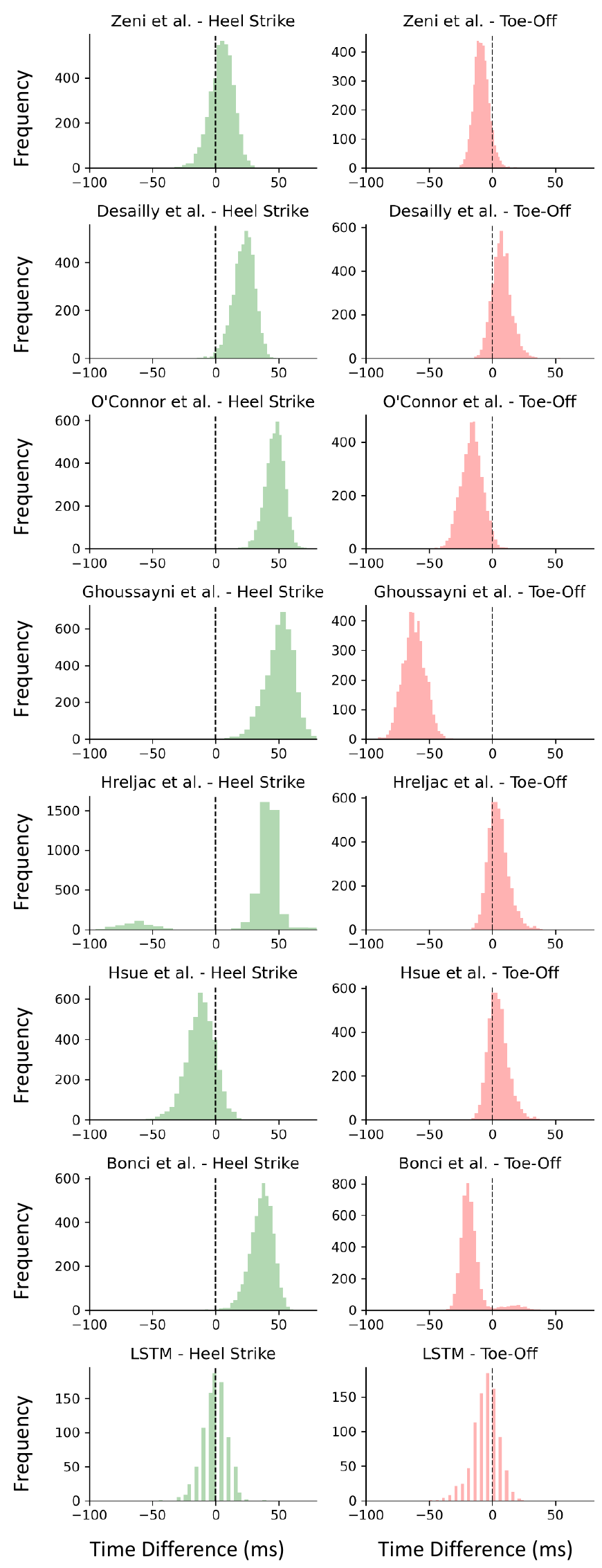}
	\caption{The histogram of time differences between the measured and calculated heel strike (left) and toe off (right) events using various kinematics-based methods and the LSTM model.
 }
\label{Histroygram}
\end{figure}

\textit{(b) LSTM Networks}

LSTM models were developed to identify gait events using the position and velocity of pelvic, heel, and toe markers. These markers included those placed on the left anterior superior iliac spine (LASIS), left posterior superior iliac spine (LPSIS), left heel (LFCC), left second metatarsal head (LFMT2), right anterior superior iliac spine (RASIS), right posterior superior iliac spine (RPSIS), right heel (RFCC), and right second metatarsal head (RFMT2). Additionally, the heel position relative to the pelvis was used as input. The model begins with a Masking layer to ignore padded values (zeros) in the input. It includes a single LSTM layer with 128 units and a 30\% dropout rate, followed by a Dense layer with 32 neurons using ReLU activation to capture high-level feature representations. Finally, a time-distributed Dense layer with a softmax activation function outputs probabilities across five nodes, corresponding to no gait event, left heel HS/TO, and right HS/TO events. Gait events were detected using a simple peak detection algorithm with a threshold of 0.01, as proposed by Lempereur et al. \cite{lempereur2020new}.

Hyperparameter optimization was conducted using Keras Tuner's RandomSearch approach. The search space included key parameters such as the number of LSTM units (32–128, in steps of 32), dropout rates (0.2–0.5, in steps of 0.1), batch size (16, 32, 64, 128), and optimizer type (Adam or RMSprop). A total of 20 trials were evaluated, each testing a unique combination of hyperparameters, with validation accuracy serving as the optimization metric. An early stopping mechanism was applied to ensure efficient tuning without overfitting. After completing the search, the best-performing hyperparameters were used to build and train the final model.

\subsection{Evaluation Protocol}
Gait event identification accuracy was evaluated by computing the temporal error (time difference) between the estimated and actual event times across both kinematics-based and machine learning-based models. The resulting discrepancies were analyzed to assess the performance of each approach in detecting key gait events.

\begin{table}[h!]
    \centering
    \caption{\TableCapWord{M}{ean} \TableCapMinor{and} \TableCapWord{S}{tandard} \TableCapWord{D}{eviation} (\TableCapWord{STD}{}) \TableCapMinor{of} \TableCapWord{T}{ime} \TableCapWord{D}{ifferences} \TableCapMinor{for} \TableCapWord{H}{eel} \TableCapWord{S}{trike} \TableCapMinor{and} \TableCapWord{T}{oe} \TableCapWord{O}{ff} \TableCapWord{E}{vents} \TableCapWord{A}{cross} \TableCapWord{K}{inematics-}\TableCapWord{B}{ased} \TableCapWord{M}{ethods} \TableCapMinor{and} \TableCapMinor{the} \TableCapWord{LSTM}{} \TableCapWord{M}{odel}.}
    \label{tab:time_diff}
    \begin{tabular}{lcc|cc}
        \toprule
        \multirow{2}{*}{Method} & \multicolumn{2}{c}{Heel Strike} & \multicolumn{2}{c}{Toe off} \\
        \cmidrule(lr){2-3} \cmidrule(lr){4-5}
        & Mean (ms) & STD (ms) & Mean (ms) & STD (ms) \\
        \midrule
        Zeni et al. & 4.78  & 9.56  & -8.66  & 6.05  \\
        Desailly et al. & 22.17 & 8.66  & 7.45   & 7.62  \\
        O'Connor et al. & 46.73 & 7.72  & -16.27 & 8.35  \\
        Ghoussayni et al. & 50.49 & 11.28 & -61.96 & 8.67  \\
        Hreljac et al. & 32.28 & 31.56 & 4.70   & 8.14  \\
        Hsue et al. & -11.86 & 11.50 & 4.70   & 8.14  \\
        Bonci et al. & 36.31 & 9.15  & -16.63 & 10.34 \\
        LSTM & -0.72 & 9.56  & -6.52  & 10.34 \\
        \bottomrule
    \end{tabular}
\end{table}

\section[RESULTS]{{\fontsize{10pt}{12pt}\selectfont RESULTS}}
Among the kinematics-based methods, the Zeni et al. approach had the smallest time difference for heel strike (4.78 $\pm$ 9.56 ms, Table \ref{tab:time_diff}), aligning most closely with the zero time difference line, followed by the methods of Hsue et al. and Desailly et al. (Fig. \ref{boxplot} and \ref{Histroygram}). In contrast, the methods of Desailly et al., O’Connor et al., Ghoussayni et al., and Bonci et al. detected heel strike later than the measured values, while the Hreljac et al. method exhibited both early and late detections. The highest variability was observed in the Hreljac et al. method (31.56 ms), indicating less consistency across trials (Fig. \ref{boxplot}).

For toe off detection, the methods of Desailly et al., Hreljac et al., and Bonci et al. had time differences close to zero, whereas the Ghoussayni et al. method detected toe off events considerably earlier (-61.96 $\pm$ 8.67 ms), suggesting a systematic bias toward early detections. O'Connor et al. and Bonci et al. also showed a tendency for earlier detections, but with smaller deviations (-16.27 ms and -16.63 ms, respectively). Variability among the methods was also evident, with some methods, such as O’Connor et al., showing lower variability (8.35 ms), while others, such as Hreljac et al., exhibited wider spreads in heel strike detection.

The LSTM-based approach demonstrated low mean time differences for both heel strike (-0.72 $\pm$ 9.56 ms) and toe off (-6.52 $\pm$ 10.34 ms), with values centered around zero. However, its performance was comparable to that of Zeni et al., which also showed relatively small deviations in heel strike detection. 

\section[DISCUSSION]{{\fontsize{10pt}{12pt}\selectfont DISCUSSION}}
In this study, we evaluated gait event detection using seven kinematics-based computational models and one LSTM model. Our results showed that the Zeni et al. method achieved the highest accuracy among the kinematics-based approaches, while other methods exhibited systematic biases or required specific adjustments for different datasets. The LSTM model performed similarly to the Zeni et al. method, providing a data-driven alternative that avoided systematic biases. These findings hold potential implications for gait analysis research and applications such as assistive exoskeletons, where accurate identification of stance and swing phases is crucial for controlling algorithms in varied environments. By enhancing the accuracy and robustness of gait event detection, these advancements could play a vital role in supporting rehabilitation efforts, improving clinical assessments, and advancing the development of assistive technologies.

Our results indicated that while certain kinematics-based methods, such as Zeni et al., provided relatively accurate gait event detection, others exhibited systematic shifts or greater variability. Kinematics-based approaches are widely used for gait analysis, particularly when force plates are unavailable. Among the evaluated methods, Zeni et al. demonstrated the smallest time difference for heel strike (4.78 ± 9.56 ms), likely due to the alignment of its reference dataset with our study population. Velocity threshold-based methods, such as those by O’Connor et al. and Ghoussayni et al., required careful tuning of thresholds across different datasets to maintain accuracy. Acceleration-based methods, including those by Hreljac et al. and Hsue et al., exhibited good predictive accuracy for horizontal components but were less reliable for vertical components, which appeared to be influenced by individual walking patterns. Interestingly, the position- and velocity-based method by Bonci et al. did not achieve the highest accuracy in our study. This may be attributed to the dataset used in its original validation. Bonci et al. initially tested their method against existing algorithms using data from healthy young adults (n = 20) and later assessed it in a small cohort of 10 individuals across five groups: older adults, individuals with chronic obstructive pulmonary disease, multiple sclerosis, Parkinson’s disease, and proximal femur fractures. In contrast, our study, which only had normative data, demonstrated that Zeni et al.’s method achieved strong accuracy without requiring extensive dataset-specific tuning.

The LSTM model demonstrated promising results, with low mean time differences and no systematic bias. Its performance was comparable to that of Zeni et al., suggesting that deep learning-based approaches can serve as viable alternatives to traditional rule-based methods. Unlike conventional approaches that rely on predefined thresholds, the LSTM model leverages its ability to learn temporal dependencies in gait kinematics, capturing complex relationships between movement phases \cite{Zhang2022lower,xia2024prediction,xiang2024integrating}. However, while the LSTM model performed well on unseen data within the same dataset, its generalizability to broader populations, particularly those with pathological gait patterns, remains uncertain. Deep learning models typically excel when evaluated on datasets similar to their training data, and further studies are required to assess their robustness across diverse movement patterns. Expanding the evaluation to individuals with gait impairments will provide valuable insights into the adaptability and potential clinical utility of LSTM-based gait event detection.

This study has several limitations. The evaluation was conducted solely on a normative dataset, limiting its applicability to populations with altered gait patterns. Future work should include individuals with conditions such as knee osteoarthritis or post-stroke impairments to assess the generalizability of the tested methods. Also, only one machine learning model (LSTM) was evaluated. Exploring alternative deep learning architectures, such as transformers, could provide further insights into model adaptability across different datasets. Lastly, cross-center validation with data from multiple research facilities would help determine the reproducibility and robustness of these approaches in diverse clinical and real-world settings.

\section[CONCLUSION]{{\fontsize{10pt}{12pt}\selectfont CONCLUSION}} 
This study assessed the performance of seven kinematics-based methods and one LSTM-based approach for gait event detection. While some kinematics-based methods, particularly Zeni et al., demonstrated strong accuracy, others exhibited systematic shifts or required dataset-specific tuning. The LSTM model showed promising results, performing comparably to the most accurate kinematics-based method without systematic bias. However, its applicability to individuals with gait impairments remains to be explored. Future research should extend these evaluations to clinical populations and explore additional machine learning techniques, such as transformer models, to enhance robustness and generalizability across diverse gait conditions and data collection settings.

\balance
\bibliographystyle{IEEEtran}
\bibliography{Ref}

\end{document}